%% file: main.tex
\documentclass[10pt,twocolumn,letterpaper]{article}

\usepackage[pagenumbers]{wacv}

\usepackage{subcaption}
\input{preamble}

\definecolor{wacvblue}{rgb}{0.21,0.49,0.74}
\usepackage[pagebackref,breaklinks,colorlinks,allcolors=wacvblue]{hyperref}

\def\wacvPaperID{817}
\def\confName{WACV}
\def\confYear{2027}

\title{AquaBEV: Monocular Underwater BEV Occupancy with 3D Sonar Supervision}

\author{Trung Tien Dong$^{1}$, Shengji Jin$^{2}$, Chen Chen$^{3}$, Yi Sheng$^{2}$ and Xiaomin Lin$^{1}$\\
$^{1}$ERA Lab, University of South Florida, Tampa, Florida, USA.\\ 
$^{2}$YES Lab, University of South Florida, Tampa, Florida, USA.\\
$^{3}$University of Central Florida, Orlando, Florida, USA.\\
{\tt\small dongt@usf.edu \quad xlin2@usf.edu}
}

\begin{document}
\maketitle

\input{sec/0_abstract}
\input{sec/1_intro}

\input{sec/2_related}
\input{sec/3_method}
\input{sec/4_experiments}
\input{sec/5_discussion}
\input{sec/6_conclusion}

{\small
\bibliographystyle{ieeenat_fullname}
\bibliography{main}
}

\end{document}

%% file: preamble.tex
\usepackage{amsmath,amssymb}
\usepackage{booktabs}
\usepackage{multirow}
\usepackage{graphicx}
\usepackage{subcaption}
\usepackage{xcolor}
\usepackage{enumitem}

%% file: sec/0_abstract.tex
% \begin{abstract}

% Autonomous underwater robots are increasingly being used for exploration, monitoring, and inspection, where a reliable understanding of the surrounding free and occupied space is essential for safe navigation. However, underwater perception remains challenging due to limited visibility, strong appearance variation, sparse geometric observations, and uncertain relationships between heterogeneous sensors. Bird's eye view (BEV) occupancy provides a compact spatial representation that can support planning and navigation, but learning such a representation directly from underwater imagery remains largely unexplored. We introduce \textbf{AquaBEV}, a monocular underwater occupancy model that predicts local BEV occupancy from a single RGB image while using paired 3D imaging sonar only as geometric supervision during training. AquaBEV maps visual features into a calibration free polar representation and applies causal decoding along the range dimension before reconstructing the prediction in Cartesian BEV coordinates. We further establish a controlled underwater occupancy benchmark by adapting representative occupancy methods to the same RGB to sonar task under a unified training and evaluation protocol. AquaBEV achieves the strongest performance  with consistent improvements in observed and distant occupancy prediction. Ablation experiments further show that the primary gain comes from preserving the ordered structure of range rather than from polar coordinates alone or increasingly detailed geometric constraints.

\begin{abstract}
Autonomous underwater robots are widely used for exploration, monitoring, and inspection, where safe navigation depends on understanding the surrounding free and occupied space. Bird's eye view (BEV) occupancy provides such a representation, but predicting it from a single underwater RGB image is difficult due to limited, unreliable geometric cues from appearance alone. 3D imaging sonar offers complementary geometric measurements to supervise this task.

We introduce \textbf{\textit{AquaBEV}}, a monocular underwater occupancy model that predicts local BEV occupancy from a single RGB image, using paired 3D imaging sonar as geometric supervision during training. \textbf{\textit{AquaBEV}} maps visual features into a calibration free polar representation and applies causal decoding along the range dimension before reconstructing the prediction in Cartesian BEV coordinates. A controlled underwater occupancy benchmark was established, adapting representative occupancy methods to the same RGB to sonar task under a unified protocol. \textbf{\textit{AquaBEV}} achieves $31.4$ Visible IoU and $38.$6 Observed IoU, $4.0\%$ and $4.3\%$ relative improvements over the strongest transferred baseline.
\end{abstract}

% \begin{abstract}

% Autonomous underwater robots are widely used for exploration, monitoring, and inspection, where safe navigation depends on understanding the surrounding free and occupied space. Reliable perception is therefore essential for safe navigation because it recovers the spatial structure needed for planning and collision avoidance. Bird's eye view (BEV) occupancy provides such a representation, but predicting it from a single underwater RGB image is difficult because visual appearance alone provides limited and unreliable geometric information. 3D imaging sonar offers complementary geometric measurements that can be used to supervise underwater BEV occupancy prediction. We introduce \textbf{AquaBEV}, a monocular underwater occupancy model that predicts local BEV occupancy from a single RGB image while using paired 3D imaging sonar as geometric supervision during training. AquaBEV maps visual features into a calibration free polar representation and applies causal decoding along the range dimension before reconstructing the prediction in Cartesian BEV coordinates. We further establish a controlled underwater occupancy benchmark by adapting representative occupancy methods to the same RGB to sonar task under a unified training and evaluation protocol. AquaBEV achieves $31.4$ Visible IoU and $38.6$ Observed IoU, corresponding to relative improvements of $4.0\%$ and $4.3\%$, respectively, over the strongest transferred occupancy baseline. 

% \end{abstract}

%% file: sec/1_intro.tex
\section{Introduction}
\label{sec:introduction}
Underwater environments remain among the most challenging domains for autonomous robotic operation, with applications spanning exploration, ecological monitoring, and infrastructure inspection~\cite{mccammon2026coralhotspots,yuan2023marine,nauert2023inspection}. Recent deployments demonstrate increasingly capable autonomy, including autonomous mapping of coral reef biodiversity hotspots~\cite{mccammon2026coralhotspots} and cooperative surface underwater vehicle operation~\cite{xu2026cooperate}. However, these advances hinge on reliable navigation in unstructured, GPS denied environments with degraded visual observations~\cite{zhang2022visualslam}. 

% As missions become more autonomous, perceiving the environment well enough to determine where a robot can safely move becomes increasingly important. Much of this perception effort, however, has focused on detecting objects of interest, such as coral structures or marine life~\cite{mccammon2026coralhotspots}.

\begin{figure}[t]
    \centering
    \includegraphics[width=1.0\columnwidth, trim=2cm 9.5cm 4.4cm 0, clip]{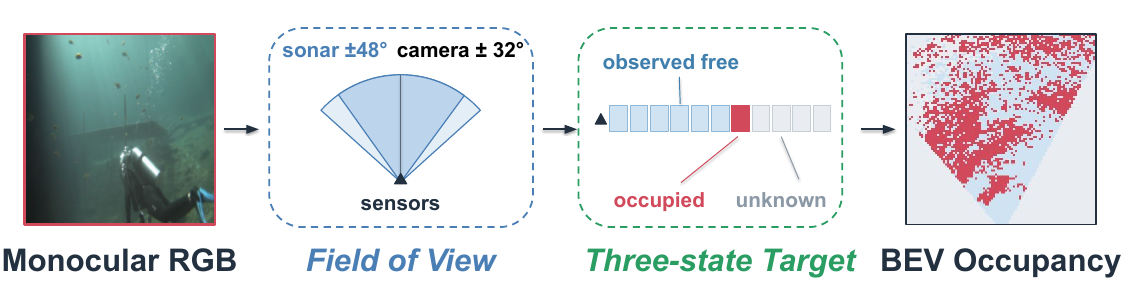}
    \caption{A single RGB image is used to predict local BEV occupancy.}
    \label{fig:task}
    \vspace{-5mm}
    
\end{figure}

\begin{figure*}[t]
    \centering
    \includegraphics[width=1.0\textwidth, trim= 2 8.86cm 30mm 2mm, clip]{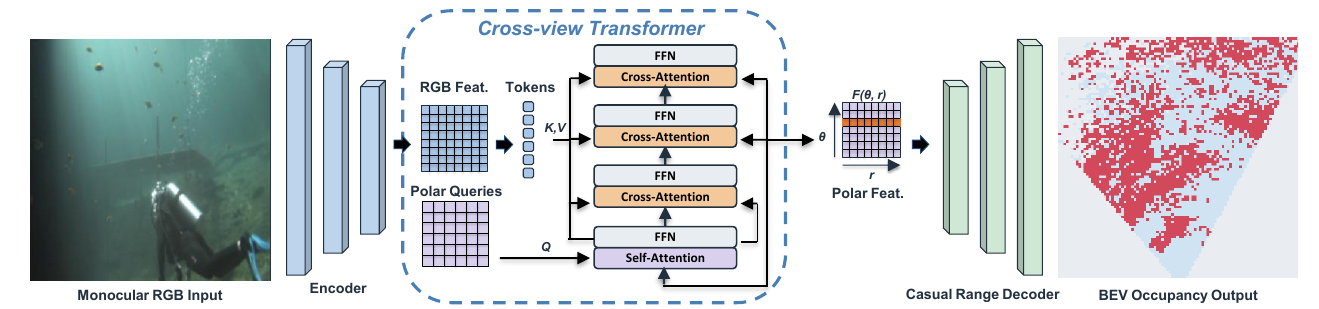}
    \caption{\textbf{Overview of AquaBEV.}
    AquaBEV encodes a single RGB image into a learned polar representation ($\theta$, r), applies causal reasoning along the range dimension, and differentiably transforms the resulting polar occupancy prediction into Cartesian BEV coordinates (x, y). }
    \label{fig:aquabev}
\end{figure*}

% Reliable navigation requires more than recognizing objects of interest. An underwater robot must also understand the spatial arrangement of surrounding structures, distinguish occupied and navigable regions, and maintain this information as it moves through the environment. Recent underwater autonomy frameworks have therefore begun incorporating occupancy maps as a shared representation between perception and navigation. DREAM~\cite{wu2025dream} constructs a persistent BEV occupancy map from visual observations to support autonomous exploration and environmental monitoring, while CORAL~\cite{wu2026coral} similarly uses a BEV occupancy representation as spatial memory for hierarchical planning and collision avoidance. These systems demonstrate the value of a compact ground-plane representation for underwater navigation. Among possible occupancy formats, BEV occupancy is a natural fit for this purpose: unlike voxel or point-based representations, it organizes local free, occupied, and unknown space in a single ground-plane coordinate frame that can be directly consumed by downstream navigation and planning \cite{dong2026post}.

Reliable navigation requires a robot to understand the spatial arrangement of surrounding structures and distinguish occupied from navigable space. Recent frameworks such as DREAM~\cite{wu2025dream} and CORAL~\cite{wu2026coral} address this by maintaining a persistent BEV occupancy map as a shared representation between perception, reasoning and navigation. Unlike voxel or point based alternatives, Bird's eye view (BEV) occupancy organizes free, occupied, and unknown space in a single ground plane frame that can be directly consumed by downstream planning~\cite{dong2026post}.

Constructing an accurate BEV occupancy underwater is extremely challenging. Cameras provide rich visual and semantic information but are strongly affected by attenuation, scattering, illumination variation, and limited visibility. Acoustic sensors provide complementary geometric observations but are typically sparse and  difficult to fuse with camera due to different sensing characteristics. Current systems sidestep this by relying on monocular depth estimation alone, CORAL~\cite{wu2026coral} being a representative example, but the resulting occupancy maps inherit the limitations of the depth estimate, which degrades under changing visual conditions.

This motivates learning BEV occupancy directly from visual observations, without depth as an intermediate step. Using both sensors at inference increases hardware and integration complexity. Instead, sonar provides geometric supervision during training while the model predicts occupancy from RGB alone at inference. The architecture therefore captures sonar’s ordered range structure during learning, as shown in Fig. \ref{fig:task}.

%At deployment, however, requiring both sensors increases sensing, calibration, and integration requirements. Learning from sonar during training while predicting occupancy from RGB alone at inference offers a way to transfer geometric information from the acoustic modality into a monocular perception model. Realizing this transfer, however, requires an occupancy architecture matched to sonar's native range-resolved geometry rather than one borrowed directly from ordinary above water perception.

% Learning this mapping is not straightforward. Modern BEV and occupancy prediction methods have largely been developed for terrestrial perception, where large annotated datasets, structured environments, and accurately characterized camera geometry are commonly available. Representative approaches use learned BEV queries~\cite{li2022bevformer}, dense monocular scene completion~\cite{cao2022monoscene}, voxel representations~\cite{li2023voxformer,zhang2023occformer}, and sparse geometric primitives~\cite{huang2024gaussianformer}. These methods provide useful foundations, but it remains unclear how their underlying representations transfer when visual observations are less stable, geometric supervision is sparse, and camera and sonar geometry may be uncertain in underwater scenario. Moreover, none were designed around the ordered, range-resolved structure that sonar-derived supervision provides. Therefore, a controlled underwater benchmark is necessary to determine which occupancy modeling principles remain effective in this setting.

Existing BEV and occupancy methods provide several possible foundations, including learned BEV queries~\cite{li2022bevformer}, dense monocular scene completion~\cite{cao2022monoscene}, voxel representations~\cite{li2023voxformer,zhang2023occformer}, and sparse geometric primitives~\cite{huang2024gaussianformer}. However, these methods were designed for terrestrial settings and do not explicitly model range ordered sonar supervision. Their transfer to underwater imagery, sparse acoustic targets, and uncertain sensor geometry therefore remains unclear. We investigate this gap through a controlled benchmark under a unified evaluation protocol.

Beyond benchmarking existing methods, we design AquaBEV to reflect the native organization of sonar measurements, as shown in Fig. \ref{fig:aquabev}. Imaging sonar records observations by azimuth and range, giving each azimuth a natural near to far order. Rather than directly decoding occupancy in Cartesian space, AquaBEV constructs a learned polar representation with causal range decoding to model the ordered range dimension. The predicted polar occupancy is then differentiably transformed into Cartesian BEV coordinates for supervision and evaluation. To the best of our knowledge, this is the first work to study monocular BEV occupancy in the underwater domain. We introduce our contributions as follows:

\begin{itemize}
    \item  We formulate monocular underwater BEV occupancy using 3D imaging sonar as training supervision and establish the first controlled benchmark for this setting. We adapt representative occupancy methods under a unified training and evaluation protocol.

    \item We introduce \textbf{AquaBEV}, the first monocular BEV occupancy model designed for underwater perception. AquaBEV uses a calibration free polar representation with causal range decoding and outperforms the strongest transferred baseline by $4.0\%$ in Visible IoU and $4.3\%$ in Observed IoU.
\end{itemize}

%% file: sec/2_related.tex
\section{Related Work}
\label{sec:related}

\subsection{Underwater Navigation and Spatial Mapping}

Autonomous underwater navigation is challenging because GPS signals do not propagate underwater. Without GPS, AUV navigation relies on inertial estimates, acoustic positioning, geophysical information, and SLAM~\cite{paull2014auv}, with visual odometry and visual SLAM further reducing drift~\cite{eustice2008visual,zhang2022visualslam}. Perception has expanded these capabilities through adaptive exploration~\cite{girdhar2014autonomous}, active visual SLAM~\cite{kim2015active}, uncertainty aware inspection planning~\cite{chaves2016opportunistic}, and autonomous mapping of biodiversity hotspots~\cite{mccammon2026coralhotspots}. Learned caveline detection~\cite{yu2023caveline}, semantic visual servoing~\cite{abdullah2024caveseg}, and integrated perception and control~\cite{gupta2025cavepi} further connect visual understanding to underwater navigation.

Recent systems also maintain persistent spatial representations for planning. DREAM combines perception and spatial reasoning for underwater monitoring~\cite{wu2025dream}, while CORAL uses occupancy for local planning and collision avoidance~\cite{wu2026coral}. However, these representations generally rely on SLAM, geometric sensing, or estimated depth. Dense occupancy prediction directly from monocular underwater imagery remains largely unexplored.

\subsection{BEV and Occupancy Prediction}

Terrestrial BEV methods transform image features using explicit geometry, as in Lift Splat Shoot~\cite{philion2020lss}, or learned queries, as in BEVFormer~\cite{li2022bevformer}. Dense occupancy methods extend this idea through monocular scene completion~\cite{cao2022monoscene}, sparse voxel queries~\cite{li2023voxformer}, and tri planar representations~\cite{huang2023tpvformer}. To reduce volumetric cost, SparseOcc uses sparse representations~\cite{tang2024sparseocc}, COTR compresses volumetric features~\cite{ma2024cotr}, and GaussianFormer models occupancy with semantic Gaussian primitives~\cite{huang2024gaussianformer,huang2025gaussianformer2}. SelfOcc instead learns geometric structure from video based rendering supervision~\cite{huang2024selfocc}.

These methods balance geometric structure, representational density, and efficiency~\cite{huang2023tpvformer,tang2024sparseocc,huang2024gaussianformer}. However, they were developed for terrestrial imagery, established camera geometry, and large datasets. Their transfer to degraded underwater imagery, sparse sonar supervision, and uncertain sensor geometry remains unclear. Cartesian representations are also poorly matched to supervision produced by a radial sensor.

\subsection{Polar and Range Structured Representations}

Polar and cylindrical representations preserve the radial organization of bearing and range measurements. PolarNet addresses range dependent LiDAR density through polar BEV partitioning~\cite{zhang2020polarnet}, Panoptic PolarNet extends this representation to panoptic segmentation~\cite{zhou2021panopticpolarnet}, and Cylinder3D preserves three dimensional structure through cylindrical partitioning~\cite{zhu2021cylinder3d}.

Because angular cells cover larger physical regions with increasing range, polar grids introduce scale and feature distortions. PolarStream applies range aware feature correction~\cite{chen2021polarstream}, PolarFormer uses multiscale polar features~\cite{jiang2023polarformer}, and PARTNER realigns features according to range and heading~\cite{nie2023partner}. PVP addresses polar occupancy distortion and cross modal misalignment~\cite{xue2025pvp}, while range conditioned convolution~\cite{bewley2021rangeconditioned} and RangeFormer~\cite{kong2023rangeformer} adapt processing to range view geometry.

These methods primarily operate directly on LiDAR measurements for segmentation, detection, or occupancy prediction~\cite{zhang2020polarnet,chen2021polarstream,nie2023partner,xue2025pvp,kong2023rangeformer}. AquaBEV instead uses polar space as an intermediate representation for predicting sonar supervised occupancy from a perspective RGB image.

%% file: sec/3_method.tex
\section{Methodology}
\label{sec:method}

\subsection{Task Formulation and Sonar Supervision}
\label{sec:task}

Given an underwater RGB image ($I_t\in\mathbb{R}^{H\times W\times3}$), AquaBEV predicts a local occupancy map in the sonar coordinate frame:
\begin{equation}
f_{\phi}(I_t)=\hat{B}_t.
\end{equation}
During training, $I_t$ is paired with 3D imaging sonar returns
\begin{equation}
    \mathcal{P}*t={(x_i,y_i,z_i)}*{i=1}^{N},
\end{equation}
which are used only to construct supervision. Sonar measurements are not provided to the network at inference.

Sonar returns within a short temporal window are expressed in the reference frame at time (t) and projected onto the horizontal plane:
\begin{equation}
(x_i,y_i,z_i)\mapsto(x_i,y_i).
\end{equation}
Each projected return marks an occupied cell. Under the first return assumption, cells traversed by the corresponding ray from the sonar origin are marked as observed free. All remaining cells are unknown because the sonar provides no evidence that they are free or occupied.

We separate binary occupancy from observation validity. Let $(Y_t(x,y)\in{0,1})$ denote free or occupied space and define
\begin{equation}
M_t(x,y)=
\begin{cases}
1, & (x,y)\text{ is acoustically observed},\cr
0, & \text{otherwise}.
\end{cases}
\end{equation}
Unknown cells therefore have $(M_t(x,y)=0)$ and are excluded from training. Figure~\ref{fig:task} summarizes this training and inference formulation.

\subsection{AquaBEV Overview}
\label{sec:aquabev_overview}

As shown in Fig.~\ref{fig:aquabev}, AquaBEV predicts Cartesian occupancy through an intermediate polar representation. A visual encoder first extracts image features. Learned polar queries then aggregate these features into a latent field indexed by azimuth and range. A causal decoder processes the field from near to far, after which a differentiable transformation maps the polar prediction into Cartesian BEV:
\begin{equation}
I_t
\rightarrow F_{\mathrm{img}}
\rightarrow F_{\mathrm{polar}}
\rightarrow H_{\mathrm{polar}}
\rightarrow \hat{B}_t.
\end{equation}

AquaBEV does not consume explicit camera to sonar extrinsic parameters. Instead, it learns the mapping from image features to the sonar aligned occupancy frame from paired training supervision. The formulation therefore avoids explicit extrinsic calibration, although the learned mapping remains tied to the sensor configuration represented during training.

\subsection{Cross View Transformer}
\label{sec:polar_representation}

% AquaBEV uses a cross view transformer to map perspective image features into a polar latent representation. A ConvNeXt B encoder extracts visual features from $I_t$. These features are flattened into image tokens and projected to a latent dimension (C). AquaBEV defines a dense polar latent field
% \begin{equation}
% F_{\mathrm{polar}}
% \in\mathbb{R}^{N_{\theta}\times N_r\times C},
% \end{equation}
% where $N_{\theta}$ and $N_r$ are the numbers of azimuth and range bins. We use $(N_{\theta}=N_r=96)$, covering azimuths from $-48^{\circ}$ to $48^{\circ}$ and ranges from 0.4 m to 10 m.

% Each polar location has a learned query that aggregates information from the image tokens through cross attention. Range positional encoding distinguishes features associated with different distances. This produces a dense polar field directly from RGB without geometrically projecting image pixels using explicit extrinsic calibration.

% The resulting field is not a sonar range image. It is a latent representation inferred entirely from RGB, whose spatial organization is learned through sonar derived occupancy supervision.

AquaBEV uses a cross view transformer to map perspective image features into a polar latent representation. A ConvNeXt B encoder first extracts a visual feature map from $I_t$. The features are flattened and projected into image tokens
\begin{equation}
T_{\mathrm{img}}
\in
\mathbb{R}^{N_{\mathrm{img}}\times C},
\end{equation}
where $N_{\mathrm{img}}$ is the number of tokens and $C$ is the latent dimension.

The polar representation is initialized using one learned query for each azimuth and range location. We use $N_{\theta}=N_r=96$, covering azimuths from $-48^{\circ}$ to $48^{\circ}$ and ranges from $0.4$ m to $10$ m. Range positional encoding is added to distinguish locations associated with different distances.

\begin{equation}
\operatorname{CrossAttn}
\left(
Q^{(\ell-1)},
T_{\mathrm{img}},
T_{\mathrm{img}}
\right).
\end{equation}
Through repeated refinement, the transformer learns the correspondence between perspective image content and locations in the polar prediction space.

The refined queries are reshaped into the dense polar field
\begin{equation}
F_{\mathrm{polar}}
\in
\mathbb{R}^{N_{\theta}\times N_r\times C},
\end{equation}
which is passed to the causal range decoder.

\input{tables/table1.tex}

\subsection{Causal Range Decoder}
\label{sec:causal_decoder}

A conventional two dimensional decoder treats azimuth and range symmetrically. AquaBEV instead models range as an ordered dimension. For azimuth $(\theta)$, let
\begin{equation}
F_{\theta}
=
[F_{\theta,1},\ldots,F_{\theta,N_r}]
\end{equation}
denote latent features ordered from near to far. The decoder output at range (r) depends only on the current and nearer polar features:
\begin{equation}
H_{\theta,r}
=
\mathcal{D}(F_{\theta,1:r}).
\end{equation}
Within the polar decoder, this constraint requires
\begin{equation}
\frac{\partial H_{\theta,r}}
{\partial F_{\theta,r'}}
=0,
\qquad r'>r.
\label{eq:causal}
\end{equation}

We implement $\mathcal{D}$ using four gated causal convolution blocks along range, with dilation rates $(d\in{1,2,4,8})$. Lightweight angular interaction additionally exchanges information between neighboring azimuth bins at the same range. Azimuth therefore captures relationships between nearby viewing directions, while range preserves directional ordering from the sensor origin.

The decoder produces
\begin{equation}
H_{\mathrm{polar}}
\in\mathbb{R}^{N_{\theta}\times N_r\times C},
\end{equation}
which is converted into polar occupancy logits by the prediction head.

\subsection{Polar to Cartesian Reconstruction}
\label{sec:polar_warp}

Let
\begin{equation}
L^{\mathrm{polar}}
\in\mathbb{R}^{N_{\theta}\times N_r}
\end{equation}
denote the predicted polar logits. A Cartesian BEV location (x,y) corresponds to
\begin{equation}
r(x,y)=\sqrt{x^2+y^2},
\qquad
\theta(x,y)=\operatorname{atan2}(y,x).
\end{equation}
We transform the prediction into Cartesian coordinates using differentiable sampling:
\begin{equation}
L^{\mathrm{BEV}}(x,y)
=
\mathcal{W}
\left(
L^{\mathrm{polar}},
\theta(x,y),
r(x,y)
\right).
\label{eq:polarwarp}
\end{equation}

Gradients from the Cartesian objective propagate through $(\mathcal{W})$ into the polar decoder. AquaBEV is therefore optimized for the final BEV task without requiring independently rasterized polar labels. 

\subsection{Training Objective}
\label{sec:objective}

The Cartesian logits are supervised only at acoustically observed cells. Given a per cell binary occupancy loss $(\ell_{\mathrm{occ}})$, the objective is
\begin{equation}
\mathcal{L}*{\mathrm{occ}}
=
\frac{
\sum*{x,y}
M_t(x,y)
\ell_{\mathrm{occ}}
\left(
L^{\mathrm{BEV}}(x,y),
Y_t(x,y)
\right)
}{
\sum_{x,y}M_t(x,y)
}.
\label{eq:occloss}
\end{equation}

The complete model is trained through this Cartesian occupancy objective. No polar labels or sonar measurements are provided as network inputs. At inference, AquaBEV maps a single RGB image directly to local Cartesian BEV occupancy.

%% file: tables/table1.tex
% ==========================================================================
% TABLE 1: Underwater BEV occupancy benchmark
% ==========================================================================

\begin{table*}[t]
\centering
\normalsize

\renewcommand{\arraystretch}{1.45}
\resizebox{\textwidth}{!}{
\begin{tabular}{@{}l|cc|cc|cccc|cc@{}}
\toprule

\multirow{2}{*}{\textbf{Method}}
& \multicolumn{2}{c|}{\textbf{Design}}
& \multicolumn{2}{c|}{\textbf{Occupancy}}
& \multicolumn{4}{c|}{\textbf{Range}}
& \multicolumn{2}{c}{\textbf{Geometry}} \\

\cline{2-11}

&
\textbf{Representation}
&
\textbf{Geometry}
&
\textbf{Visible} $\uparrow$
&
\textbf{Observed} $\uparrow$
&
\textbf{Near} $\uparrow$
&
\textbf{Mid} $\uparrow$
&
\textbf{Far} $\uparrow$
&
\textbf{R Macro} $\uparrow$
&
\textbf{bF1@2} $\uparrow$
&
\textbf{Chamfer} $\downarrow$
\\

\midrule

SurroundOcc~\cite{wei2023surroundocc}
& Dense voxel
& Projection based
& 29.8
& 34.9
& 38.9
& \underline{40.6}
& 18.9
& 32.8
& \textbf{48.4}
& \underline{0.355}
\\

BEVFormer~\cite{li2022bevformer}
& BEV queries
& Projection based
& 28.7
& 34.6
& 39.7
& 39.5
& 18.4
& 32.6
& 45.1
& 0.398
\\

VoxFormer~\cite{li2023voxformer}
& Sparse voxel
& Projection based
& 29.2
& 35.7
& 38.8
& 39.6
& 19.7
& 32.7
& 47.0
& 0.371
\\

MonoScene~\cite{cao2022monoscene}
& Dense voxel
& Projection based
& 29.8
& 35.8
& 40.8
& 39.7
& 20.3
& 33.6
& 45.4
& 0.381
\\

TPVFormer~\cite{huang2023tpvformer}
& Tri perspective
& Projection based
& 29.0
& 34.2
& 37.4
& 39.3
& 18.9
& 31.8
& \underline{47.2}
& 0.382
\\

OccFormer~\cite{zhang2023occformer}
& Dual path voxel
& Projection based
& 28.8
& 36.0
& 38.5
& 40.0
& 19.6
& 32.7
& 46.4
& 0.378
\\

GaussianFormer$^{\dagger}$~\cite{huang2024gaussianformer}
& Sparse Gaussian
& Calibration free
& \underline{30.2}
& \underline{37.0}
& \underline{41.9}
& 39.6
& \underline{21.4}
& \underline{35.0}
& 41.3
& 0.412
\\

\midrule

\textbf{AquaBEV (ours)}
& \textbf{Polar occupancy}
& \textbf{Calibration free}
& \textbf{31.4}
& \textbf{38.6}
& \textbf{43.0}
& \textbf{42.1}
& \textbf{22.2}
& \textbf{35.8}
& 45.9
& \textbf{0.340}
\\

\bottomrule
\end{tabular}
}
\caption{ \textbf{Comparison of monocular underwater BEV occupancy methods on the proposed benchmark.} All methods use the same RGB input, three state sonar supervision, data split, ConvNeXt B backbone, training schedule, and evaluation protocol (\textbf{BEST}/ \underline{SECOND BEST}).
}
\label{tab:main_benchmark}
\vspace{-2mm}
\end{table*}

%% file: sec/4_experiments.tex
\section{Experiments}
\label{sec:experiments}

\input{tables/table2A.tex}

\input{tables/table2B.tex}

\subsection{Dataset and Evaluation Protocol}
\label{sec:dataset}

We evaluate monocular underwater BEV occupancy using paired RGB and 3D imaging sonar collected across seven underwater sessions \cite{dong2026uscenes}. The complete dataset contains 110 scenes and approximately 96,000 paired observations. We use sessions $\{2,3,6,7,8\}$ for training, session $5$ for validation, and session $4$ for testing. The split is fixed for all experiments so that every method is evaluated on the same unseen session.

For every RGB frame, the corresponding occupancy target is constructed from accumulated 3D imaging sonar measurements using the three state formulation described in Sec.~\ref{sec:task}. Each target uses 16,384 sonar points and contains occupied, observed free, and unknown cells. The same target generation procedure is used for all benchmark models and AquaBEV. At inference, every method receives only a single RGB image. Table~\ref{tab:main_benchmark} reports results under this fixed protocol.

We evaluate occupancy using complementary overlap, range, and geometric metrics. \textbf{Visible IoU} measures occupied cell overlap across the camera visible portion of the BEV. Because this region can contain cells without acoustic observation, predictions in unknown cells are counted as false positives. \textbf{Observed IoU} instead restricts evaluation to occupied and observed free cells using the observation mask $(M_t)$. This metric excludes unknown cells and therefore directly measures performance over supervised space.

To evaluate occupancy as a function of forward distance, we divide the visible BEV into three bands along the forward $x$ axis. Near covers $x\in[0.5,3.0)$ m, Mid covers $x\in[3.0,6.0)$ m, and Far covers $x\in[6.0,10.0)$ m. Each band is intersected with the camera visible field of view before evaluation. These intervals measure forward distance rather than polar slant range, which avoids conflating the evaluation bands with the radial coordinate used internally by AquaBEV.

We report IoU independently within each band and define \textbf{Range Macro} as the equally weighted mean
\begin{equation}
    \mathrm{RMacro}
    =
    \frac{
        \mathrm{IoU}_{\mathrm{near}}
        +
        \mathrm{IoU}_{\mathrm{mid}}
        +
        \mathrm{IoU}_{\mathrm{far}}
    }{3}.
\end{equation}
The metric therefore gives equal importance to near, mid, and far occupancy despite their different spatial widths. The small interval $x\in[0.4,0.5)$ m is excluded from the three distance bands but remains included in Visible and Observed IoU.

Occupancy overlap alone does not fully describe geometric quality. We therefore report \textbf{boundary F1} with a tolerance of two BEV cells and \textbf{Chamfer distance} between predicted and target occupancy boundaries. Higher IoU and boundary F1 indicate better performance, while lower Chamfer distance indicates better geometric agreement.

All models use ConvNeXt B with a latent dimension of (192) and are trained for (30) epochs with batch size (24). We use the AdamW optimizer with initial learning rate of $(3\times10^{-4})$, weight decay of $(1\times10^{-4})$, cosine scheduling, BEV flipping, and MixUp. Checkpoints are selected using validation Visible IoU, and predictions are thresholded at (0.5). Transferred methods use seed (0), while AquaBEV and its primary ablations report the mean and sample standard deviation over three seeds.

\subsection{Comparison with Existing Occupancy Methods}
\label{sec:benchmark_results}

Table~\ref{tab:main_benchmark} compares AquaBEV with seven representative occupancy methods under the unified training and evaluation protocol. AquaBEV achieves $31.4$ Visible IoU and $38.6$ Observed IoU, compared with $30.2$ and $37.0$ for GaussianFormer, the strongest transferred baseline on these metrics.

Across forward distance, AquaBEV obtains $43.0$ Near IoU, $42.1$ Mid IoU, and $22.2$ Far IoU. It achieves the strongest overall Range Macro among the evaluated methods. 

AquaBEV also achieves the lowest Chamfer distance at $0.340$ m. SurroundOcc obtains the highest boundary F1 at $48.4$, compared with $45.9$ for AquaBEV. These results show that AquaBEV achieves the strongest overall occupancy performance while remaining competitive in geometric quality.

\begin{figure*}[t]
    \centering
    \includegraphics[width=1.0\textwidth, trim= 0 0 0 0, clip]{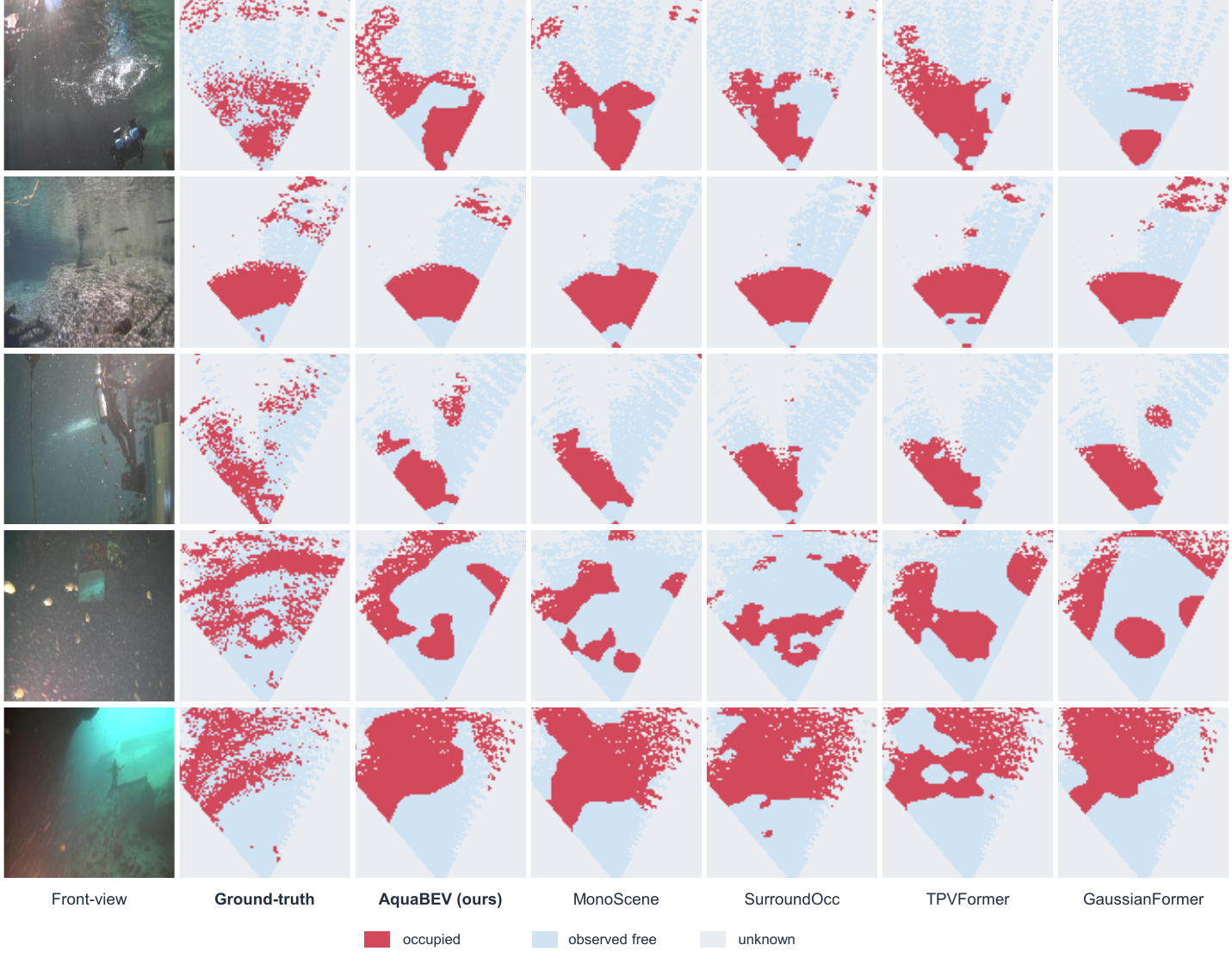}
    \caption{\textbf{Qualitative comparison of monocular underwater BEV occupancy.}}
    \label{fig:qualitative}
\end{figure*}

\subsection{AquaBEV Ablation}
\label{sec:ablation}

\textbf{Encoder selection.} Table~\ref{tab:encoder_selection} reports an encoder study. ConvNeXt B achieves the highest validation and test Visible IoU and matches the highest Near IoU. A subsequent three seed comparison also favors ConvNeXt B over ResNet 50, with $(32.6\pm0.34)$ versus $(31.7\pm0.82)$ Visible IoU. We therefore use ConvNeXt B for all final experiments.

\textbf{Representation and range decoder.} Table~\ref{tab:aquabev_ablation} shows that changing from Cartesian to polar decoding provides a modest improvement, increasing Visible IoU from (30.8) to (31.3). Causal range decoding produces the main reproducible gain, increasing Observed IoU from (37.5) to (38.6), with a paired three seed improvement of $(1.10\pm0.20)$ points. Survival supervision, metric angular coupling, and full range context provide no consistent improvement. AquaBEV also achieves the lowest Chamfer distance at (0.340) m.

\subsection{Qualitative Results}
\label{sec:qualitative}

Figure~\ref{fig:qualitative} compares the Cartesian occupancy predictions of AquaBEV and the transferred baselines with the sonar derived ground truth. AquaBEV produces more continuous occupied regions and better preserves their spatial extent across near and distant ranges. The improvements are most visible for elongated structures and separated occupancy regions, where competing methods often produce fragmented predictions, miss portions of the structure, or introduce isolated false positives. These observations are consistent with AquaBEV's higher Visible, Observed, and Far IoU.

Challenges remain when structures are visually indistinct, weakly illuminated, or only sparsely observed by sonar. In these regions, AquaBEV may underestimate occupied extent, merge nearby structures, or fail to recover distant occupancy. Such errors reflect the inherent ambiguity of inferring sonar supervised geometry from a single RGB image, particularly when appearance provides limited evidence of distance or occluded structure.

\begin{figure*}[t]
    \centering
    \includegraphics[width=1.0\textwidth, trim= 0 0 0 0, clip]{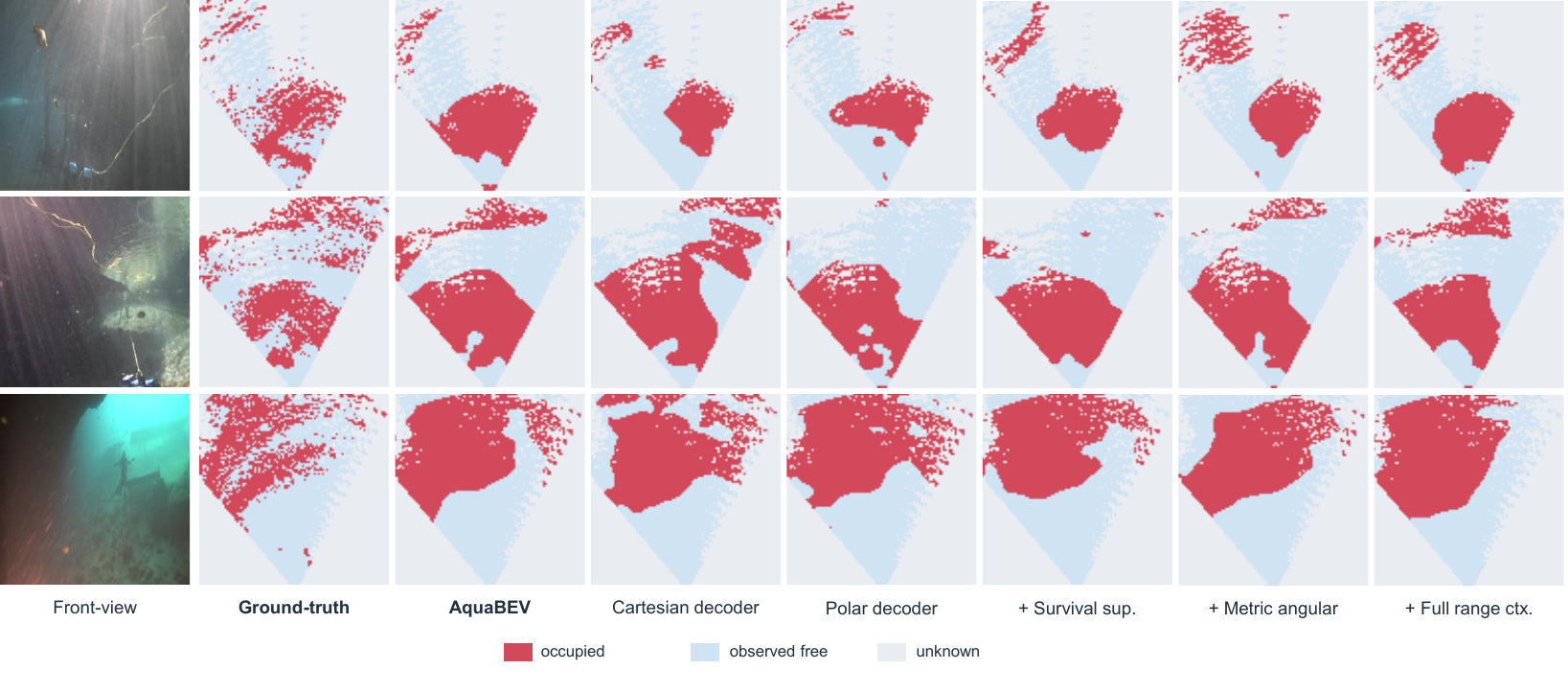}
    \caption{\textbf{Ablation comparison of Casual Range Decoder.}}
    \label{fig:ablation}
\end{figure*}

%% file: tables/table2A.tex
\begin{table*}[t]
\centering

\small
\setlength{\tabcolsep}{4.5pt}
\renewcommand{\arraystretch}{1.18}

\begin{tabular}{@{}lcccccccc@{}}
\toprule
\textbf{Method}
& \textbf{Vis.} $\uparrow$
& \textbf{Obs.} $\uparrow$
& \textbf{Near} $\uparrow$
& \textbf{Mid} $\uparrow$
& \textbf{Far} $\uparrow$
& \textbf{R Macro} $\uparrow$
& \textbf{bF1@2} $\uparrow$
& \textbf{Chamfer} $\downarrow$ \\
\midrule

Cartesian decoder
& 30.8
& 37.4
& 42.4
& 42.0
& 20.7
& 35.0
& 47.7
& 0.351 \\

Polar decoder
& 31.3
& 37.5
& 42.3
& 42.0
& 21.5
& 35.2
& \textbf{48.0}
& 0.342 \\

Polar + Survival supervision
& 31.3
& 38.5
& 42.6
& 42.0
& 21.7
& 35.7
& 45.8
& 0.344 \\

Polar + Metric angular coupling
& 31.0
& 38.4
& 42.5
& 42.0
& 21.5
& 35.3
& 46.2
& 0.349 \\

Polar + Full range context
& 30.9
& \textbf{38.6}
& 41.1
& 42.0
& 22.1
& 35.0
& 44.1
& 0.360 \\

\midrule

\textbf{AquaBEV}
& \textbf{31.4}
& \textbf{38.6}
& \textbf{43.0}
& \textbf{42.1}
& \textbf{22.2}
& \textbf{35.8}
& 45.9
& \textbf{0.340} \\

\bottomrule
\end{tabular}

\caption{
\textbf{Ablation of AquaBEV on occupancy and geometric quality.}
All results are averaged over three seeds.
}
\label{tab:aquabev_ablation}

\vspace{-2mm}
\end{table*}

%% file: tables/table2B.tex
\begin{table}[t]
\centering

\small
\setlength{\tabcolsep}{4.2pt}
\renewcommand{\arraystretch}{1.15}

\begin{tabular}{@{}lcccc@{}}
\toprule
\textbf{Encoder}
& \textbf{Params.}
& \textbf{Val.}
& \textbf{Vis.} $\uparrow$
& \textbf{Near} $\uparrow$ \\
\midrule

DINOv2 S
& 23.2M
& 30.6
& 28.2
& 39.0 \\

ResNet 50
& 25.6M
& 33.9
& 32.7
& \textbf{45.3} \\

ConvNeXt T
& 29.5M
& 33.5
& 30.9
& 42.6 \\

ResNet 101
& 44.6M
& 33.6
& 30.1
& 42.1 \\

ConvNeXt S
& 51.2M
& 33.5
& 29.1
& 40.1 \\

ResNet 152
& 60.3M
& 33.8
& 31.3
& 43.9 \\

DINOv2 B
& 87.3M
& 29.3
& 28.0
& 38.7 \\

\textbf{ConvNeXt B}
& 89.4M
& \textbf{34.9}
& \textbf{32.9}
& \textbf{45.3} \\

\bottomrule

\end{tabular}

\caption{
\textbf{RGB encoder ablation.}
All encoders are evaluated using the same calibration free occupancy framework, training protocol, and held out test session. The results are from one seed.}
\label{tab:encoder_selection}

\vspace{-2mm}
\end{table}

%% file: sec/5_discussion.tex
\section{Discussion}
\label{sec:discussion}

\subsection{Occupancy Transfer to Underwater Perception}
\label{sec:discussion_transfer}

The benchmark reveals a notable characteristic of underwater monocular occupancy prediction. Although the transferred methods use substantially different representations, their performance remains relatively concentrated. Visible IoU spans only $28.7$ to $30.2$ across the seven transferred methods, while Observed IoU ranges from $34.2$ to $37.0$. The narrow performance range suggests that changing the occupancy representation alone has limited impact in this setting.

A possible explanation is that the primary challenge lies in recovering reliable geometry from a single underwater image. A single underwater RGB image must infer geometry supervised by a sensing modality with substantially different visibility and measurement characteristics. Increasing the complexity of the occupancy decoder cannot recover geometric information that is weak or ambiguous in the image itself. This is particularly apparent at farther distances, where every method experiences a substantial reduction in IoU.

The benchmark also suggests that geometric assumptions should be transferred carefully. Several existing approaches were designed around environments with well characterized camera geometry and stable scene structure. AquaBEV instead learns the mapping from visual features to occupancy without using camera and sonar extrinsic parameters during feature transport. Its improvement does not establish that explicit projection is generally inferior, but it indicates that learned transport is a useful alternative when reliable cross sensor geometry is difficult to obtain.

The differences between evaluation metrics are also informative. AquaBEV achieves the strongest occupancy overlap and lowest Chamfer distance, while SurroundOcc achieves higher boundary F1. No single metric therefore captures the complete quality of an occupancy prediction. Overlap measures how much occupied structure is recovered, while boundary metrics measures emphasize local contour alignment and Chamfer measures the spatial displacement between structures. Reporting these jointly is particularly important for sparse underwater scenes.

\subsection{Why Causal Range Reasoning Helps}
\label{sec:discussion_range}

The ablation study separates two ideas that can otherwise be easily conflated: using polar coordinates and reasoning along an ordered range dimension. Polar coordinates alone provide only a small improvement over the matched Cartesian decoder. The larger and reproducible change appears only after causal range decoding is introduced. This indicates that the benefit of AquaBEV does not come simply from expressing the scene in polar coordinates, as shown in Fig. \ref{fig:ablation}.

The result can be understood from the different roles of the two polar axes. Azimuth describes neighboring viewing directions, while range describes progression away from the sensor origin. Treating both dimensions identically with conventional spatial convolution ignores this distinction. AquaBEV instead allows nearby azimuth regions to interact while imposing an ordering along range, so a prediction at a given distance is constructed without information from farther locations.

The additional ablations further isolate this effect. Metric angular coupling explicitly compensates for the increasing physical width represented by an angular cell at greater distance, yet it does not improve prediction. Likewise, extending the causal receptive field across the complete range sequence provides essentially the same Observed and Far IoU as the shorter AquaBEV decoder. The improvement therefore cannot be explained by greater context alone.

Explicit survival supervision also provides no consistent gain. This is important because it limits how AquaBEV should be interpreted. The model is not learning a strict first return sonar process. After 3D sonar measurements are projected into the horizontal BEV plane, several surfaces at different heights can occupy different distances along the same direction. A simple free region followed by one acoustic return is therefore not generally consistent with the supervision target.

Taken together, the results support a narrower conclusion: the useful sonar informed inductive bias is the \emph{ordering of range}, rather than a complete physical model of sonar propagation. This distinction allows AquaBEV to exploit structure associated with radial sensing without requiring the occupancy target to satisfy an overly restrictive acoustic model.

%% file: sec/6_conclusion.tex
\section{Conclusion}
\label{sec:conclusion}
\subsection{Summary of Findings}
We studied monocular underwater BEV occupancy using offline 3D imaging sonar supervision and established a controlled benchmark for transferring representative occupancy methods to this setting. Under a unified training and evaluation protocol, AquaBEV achieves the strongest overall occupancy performance among the evaluated methods while requiring only a single RGB image at inference.

AquaBEV combines a learned polar occupancy representation with causal reasoning along range. Our ablations show that polar coordinates alone provide only a modest improvement, while causal range decoding produces the main reproducible gain. Additional survival supervision, range conditioned angular coupling, and longer range context do not further improve performance. These results indicate that ordered range reasoning is a more useful inductive bias than detailed modeling of sonar propagation.

The current study remains limited to one underwater environment and evaluates AquaBEV only as a perception model. Sonar accumulation and projection also introduce geometric and temporal ambiguities into the supervision. Improved occupancy prediction therefore does not yet establish improved navigation performance.

\subsection{Future Work}
\label{sec}

We plan to integrate AquaBEV with existing underwater navigation frameworks such as CORAL~\cite{wu2026coral}. Closed loop experiments will evaluate whether the predicted occupancy improves path planning, collision avoidance, and navigation efficiency in practice.

We will also extend AquaBEV from binary occupancy to semantic BEV prediction by incorporating segmentation of underwater structures, terrain, vehicles, and other relevant objects. A further direction is physics informed BEV generalization across changes in water clarity, illumination, acoustic observation density, and sensor configuration. This will examine whether underwater imaging and sonar priors can improve transfer to environments and operating conditions not represented during training.